\documentclass{fcs}
\usepackage{bm}
\usepackage{multirow}
\usepackage{enumerate}
\usepackage{wrapfig}
\usepackage{graphicx}
\usepackage{hyperref}
\usepackage{subfigure}
\usepackage{amssymb}
\usepackage{pifont}
\usepackage{algorithm, algorithmic}

\usepackage[misc,geometry]{ifsym}
\usepackage{lipsum}
\usepackage{csquotes}

\newcommand{\eg}{\emph{e.g.}}

\volumn{ }
\doi{ }
\articletype{LETTER}
\copynote{{\copyright} Higher Education Press 2024}
\ratime{Received month dd, yyyy; accepted month dd, yyyy}
\email{Corresponding author: Han-Jia Ye (yehj@lamda.nju.edu.cn)}
\title{TV100: A TV Series Dataset that Pre-Trained CLIP Has Not Seen}
\author{Da-Wei Zhou$^{1,2}$, Zhi-Hong Qi$^{1,2}$, Han-Jia Ye$^{1,2}$, De-Chuan Zhan$^{1,2}$}
\address{{1\quad National Key Laboratory for Novel Software Technology, Nanjing University, Nanjing, 210023, China}\\
{2\quad School of Artificial Intelligence, Nanjing University, Nanjing, 210023, China}}

\begin{document}
\maketitle
\setcounter{page}{1}
\setlength{\baselineskip}{14pt}

\vspace{-10mm}
\section{Introduction}

\noindent In recent years, the field of deep learning has experienced remarkable growth~\cite{ning2023rf,ye2023contextualizing,chen2022learning,yang2015auxiliary,zhou2024expandable,ning2022moirepose}, leading to the emergence of large, pre-trained models such as ChatGPT~\cite{floridi2020gpt}, which demonstrates significant capability in understanding and responding to human language inputs, and DALL-E~\cite{ramesh2021zero}, which creatively generates images from textual descriptions in a zero-shot manner. Another notable innovation in this domain is CLIP~\cite{radford2021learning} (Contrastive Language-Image Pre-Training), a model that excels in representation learning by bridging multiple modalities to perform classifications, also in a zero-shot manner. CLIP, trained on a diverse array of images and natural language descriptions readily available on the internet, can interpret natural language instructions to execute a wide range of classification tasks without specific optimization for those tasks. These advanced models have shown remarkable effectiveness in various real-world applications, showcasing their potential even when not trained on task-specific data. Notably, CLIP achieved a zero-shot accuracy of 76.2\% on the ImageNet~\cite{deng2009imagenet} dataset. However, a pressing question remains within the machine learning community:
\begin{displayquote} 
	{\bf Does CLIP know everything?}
\end{displayquote}
This question is pivotal. If a model could truly understand and react to all information, the exploration of alternative models might become redundant. Nevertheless, the reality is that no model, including CLIP, possesses complete knowledge. Our world is in constant flux, with new data, objects, categories, and information emerging regularly~\cite{zhou2023class}. For instance, ChatGPT's knowledge of world events, such as political changes, is contingent upon its training data, and CLIP cannot recognize images of products released after its last update, such as the 'Apple Vision Pro' launched in 2023.

This paper focuses on identifying datasets unknown to CLIP, a task of considerable importance. Given CLIP's training on the extensive LAION dataset~\cite{schuhmann2022laion}, identifying such datasets not only facilitates the application of transfer learning for downstream tasks but also serves as a means to evaluate CLIP's ability to detect out-of-distribution or novel instances~\cite{zhou2021learning} and continual learning~\cite{zhou2024continual,sun2023pilot}. This is particularly relevant in the context of addressing the hallucination issues prevalent in large models~\cite{rawte2023survey}. To advance research in this area, we introduce a dataset of TV series released post-2021, named {\bf TV100}, to explore CLIP's performance further. The dataset, along with detailed information, is accessible at: \url{https://tv-100.github.io/}.

\begin{figure*}[t]
	\begin{center}
		\subfigure[ Data collection pipeline ]
		{\includegraphics[width=2\columnwidth]{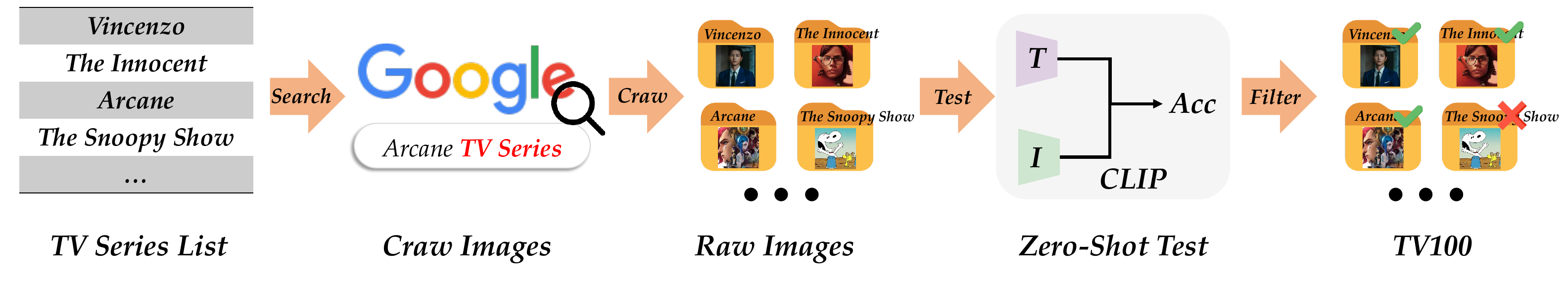}\label{figure:a}}\\ 
		\subfigure[  Country distribution]
		{\includegraphics[width=.66\columnwidth]{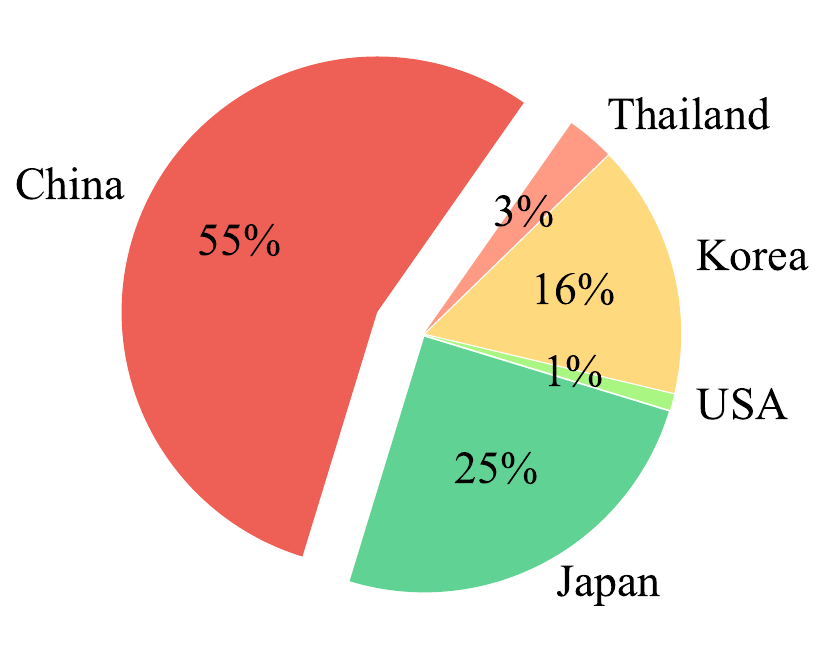}\label{figure:b}}
		\subfigure[ Instance distribution]
		{\includegraphics[width=.66\columnwidth]{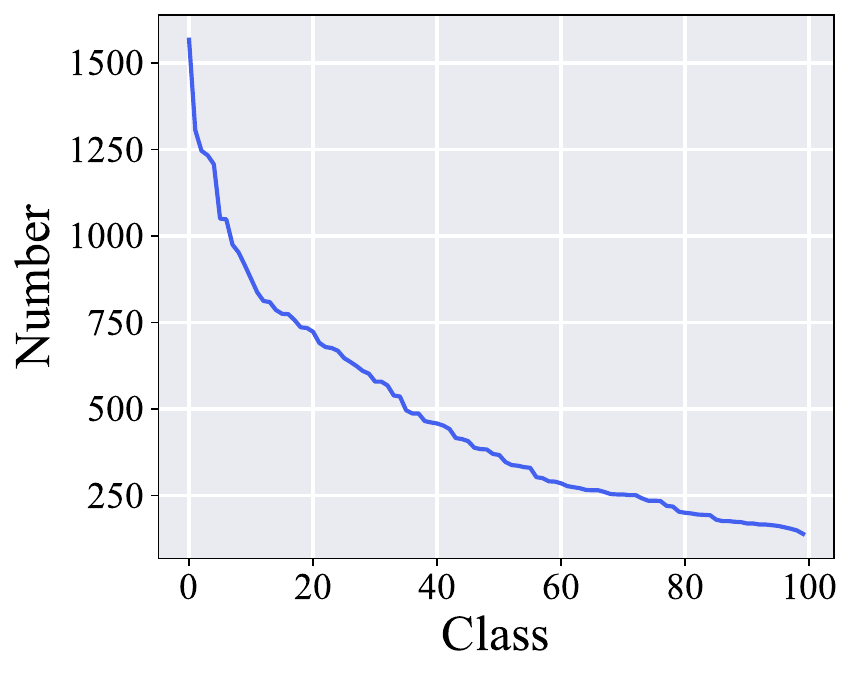}\label{figure:c}}
		\subfigure[   Performance evaluation]
		{\includegraphics[width=.66\columnwidth]{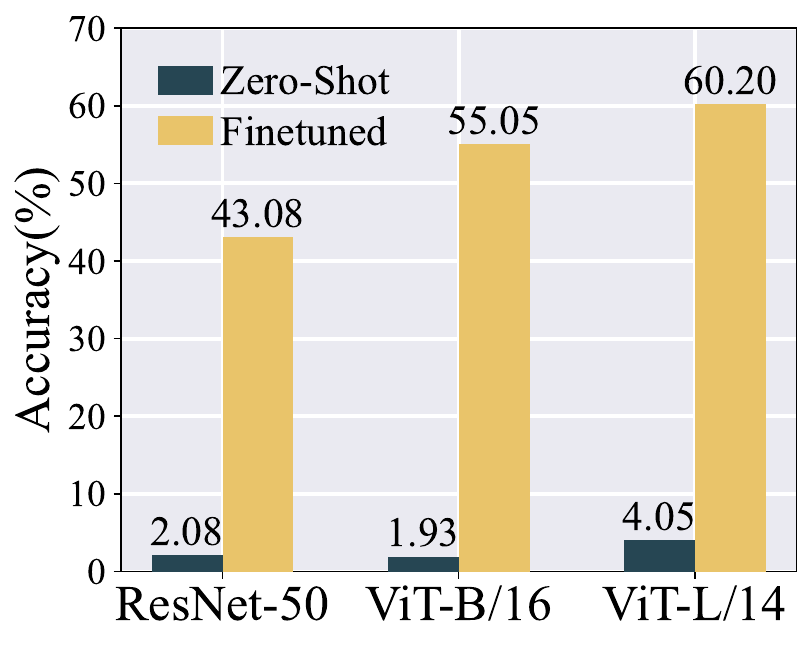}\label{figure:d}}
	\end{center}
	\caption{Detailed information about TV100, including the data collection process, the country distribution, and class distribution. It also contains an empirical evaluation of zero-shot and finetuned performance.}\label{figure1} 
\end{figure*}

\section{Dataset Construction}
\noindent\textbf{Data Construction}: We visualize the data collection process in Figure~\ref{figure:a}. Specifically, 
we manually search for TV series from IMDB and collect the items released after 2021. Afterward, we download the related images on Google by searching the keyword ``{\bf [NAME] TV Series},'' where [NAME] is the name of the TV series. The downloaded images are then processed manually to delete repeated and meaningless ones. Hence, we can get a large dataset that contains around 800 classes. However, some classes may be seen for the CLIP, \eg, ``The Snoopy Show'' (Snoopy is a famous cartoon character).
Hence, we use a pre-trained CLIP to rank the difficulty of these classes by measuring the zero-shot accuracy of each image and the text ``a photo of the TV series [CLASS].'' We choose the top-100 hard classes based on the zero-shot accuracy and construct the TV100 dataset. 

\noindent\textbf{Data Statistics}: We summarize the data statistics in Figure~\ref{figure:b} and Figure~\ref{figure:c}. Specifically, Figure~\ref{figure:b} indicates that the dataset contains TV series from countries worldwide.
 Figure~\ref{figure:c} implies that the class distribution is highly imbalanced, which is naturally suitable for scientific research on long-tailed recognition. 

\noindent\textbf{Data Evaluation}: To investigate whether a pre-trained CLIP knows these images, we conduct an experiment on its zero-shot performance and finetuned performance in Figure~\ref{figure:d}. Accordingly, we find a pre-trained CLIP cannot recognize any classes from the dataset. By contrast, if we finetune the CLIP model with the images, the performance drastically improves, indicating that the dataset is learnable and separable.


\noindent\textbf{Conclusion} The era of pre-trained models has ushered in a wealth of new insights for the machine learning community. Among the myriad of questions that arise, one of paramount importance is: 'Do pre-trained models possess comprehensive knowledge?' This paper seeks to address this crucial inquiry. In line with our objective, we have made publicly available a novel dataset comprised of images from TV series released post-2021. This dataset holds significant potential for use in various research areas, including the evaluation of novel class discovery and long-tailed learning, among others.

\Acknowledgements{This work is partially supported by National Science and Technology Major Project (2022ZD0114805),
	NSFC (62376118, 62006112, 62250069, 61921006), Collaborative Innovation Center of Novel Software
	Technology and Industrialization.}

\bibliographystyle{fcs}
\bibliography{paper}

\begin{thebibliography}{10}

\bibitem{ning2023rf}
Ning J, Xie L, Wang C, Bu~Y, Xu~F, Zhou D~W, Lu~S, Ye~B.
\newblock Rf-badge: Vital sign-based authentication via rfid tag array on
  badges.
\newblock IEEE Transactions on Mobile Computing, 2023, 22(02): 1170--1184

\bibitem{ye2023contextualizing}
Ye~H~J, Zhou D~W, Hong L, Li~Z, Wei X~S, Zhan D~C.
\newblock Contextualizing meta-learning via learning to decompose.
\newblock IEEE Transactions on Pattern Analysis and Machine Intelligence, 2023

\bibitem{chen2022learning}
Chen S, Gong C, Li~J, Yang J, Niu G, Sugiyama M.
\newblock Learning contrastive embedding in low-dimensional space.
\newblock NeurIPS, 2022, 35: 6345--6357

\bibitem{yang2015auxiliary}
Yang Y, Ye~H~J, Zhan D~C, Jiang Y.
\newblock Auxiliary information regularized machine for multiple modality
  feature learning.
\newblock In: IJCAI.
\newblock 2015

\bibitem{zhou2024expandable}
Zhou D~W, Sun H~L, Ye~H~J, Zhan D~C.
\newblock Expandable subspace ensemble for pre-trained model-based
  class-incremental learning.
\newblock arXiv preprint arXiv:2403.12030, 2024

\bibitem{ning2022moirepose}
Ning J, Xie L, Li~Y, Chen Y, Bu~Y, Ye~B, Lu~S.
\newblock Moir{\'e}pose: ultra high precision camera-to-screen pose estimation
  based on moir{\'e} pattern.
\newblock In: Proceedings of the 28th Annual International Conference on Mobile
  Computing And Networking.
\newblock 2022,  106--119

\bibitem{floridi2020gpt}
Floridi L, Chiriatti M.
\newblock Gpt-3: Its nature, scope, limits, and consequences.
\newblock Minds and Machines, 2020, 30(4): 681--694

\bibitem{ramesh2021zero}
Ramesh A, Pavlov M, Goh G, Gray S, Voss C, Radford A, Chen M, Sutskever I.
\newblock Zero-shot text-to-image generation.
\newblock In: ICML.
\newblock 2021,  8821--8831

\bibitem{radford2021learning}
Radford A, Kim J~W, Hallacy C, Ramesh A, Goh G, Agarwal S, Sastry G, Askell A,
  Mishkin P, Clark J, others .
\newblock Learning transferable visual models from natural language
  supervision.
\newblock In: ICML.
\newblock 2021,  8748--8763

\bibitem{deng2009imagenet}
Deng J, Dong W, Socher R, Li~L~J, Li~K, Fei-Fei L.
\newblock Imagenet: A large-scale hierarchical image database.
\newblock In: CVPR.
\newblock 2009,  248--255

\bibitem{zhou2023class}
Zhou D~W, Wang Q~W, Qi~Z~H, Ye~H~J, Zhan D~C, Liu Z.
\newblock Deep class-incremental learning: A survey.
\newblock arXiv preprint arXiv:2302.03648, 2023

\bibitem{schuhmann2022laion}
Schuhmann C, Beaumont R, Vencu R, Gordon C, Wightman R, Cherti M, Coombes T,
  Katta A, Mullis C, Wortsman M, others .
\newblock Laion-5b: An open large-scale dataset for training next generation
  image-text models.
\newblock NeurIPS, 2022, 35: 25278--25294

\bibitem{zhou2021learning}
Zhou D~W, Ye~H~J, Zhan D~C.
\newblock Learning placeholders for open-set recognition.
\newblock In: CVPR.
\newblock 2021,  4401--4410

\bibitem{zhou2024continual}
Zhou D~W, Sun H~L, Ning J, Ye~H~J, Zhan D~C.
\newblock Continual learning with pre-trained models: A survey.
\newblock arXiv preprint arXiv:2401.16386, 2024

\bibitem{sun2023pilot}
Sun H~L, Zhou D~W, Ye~H~J, Zhan D~C.
\newblock Pilot: A pre-trained model-based continual learning toolbox.
\newblock arXiv preprint arXiv:2309.07117, 2023

\bibitem{rawte2023survey}
Rawte V, Sheth A, Das A.
\newblock A survey of hallucination in large foundation models.
\newblock arXiv preprint arXiv:2309.05922, 2023

\end{thebibliography}

\vfill

\end{document}